\documentclass[a4paper,11pt]{article}

\usepackage[utf8]{inputenc} 
\usepackage[T1]{fontenc}    
\usepackage{hyperref}       
\usepackage{url}            
\usepackage{booktabs}       
\usepackage{amsfonts}       
\usepackage{nicefrac}       
\usepackage{microtype}      

\usepackage[authoryear,round]{natbib} 

\usepackage{amsmath}

\usepackage{mathpazo}
\usepackage{blindtext}

\usepackage[]{algorithm2e}
\usepackage{listings}
\usepackage{xcolor}
\definecolor{codebg}{rgb}{0.95,0.95,0.97}
\definecolor{codecomment}{rgb}{0.5,0.5,0.5}
\definecolor{codekeyword}{rgb}{0.8,0.25,0.2}
\definecolor{codename}{rgb}{0.0,0.3,0.9}
\definecolor{codeidentifier}{rgb}{0.15,0.15,0.15}

\usepackage[scaled]{beramono}

\makeatletter

\renewcommand{\@listI}{\itemsep=0pt}

\title{
	\vskip -45pt
	\fontsize{16}{20}\textbf{Fast Differentiable Clipping-Aware Normalization and Rescaling}
	\vskip 17pt
}
\author{
	\textbf{Jonas Rauber} \\ Department of Computer Science \\ University of Tübingen
	\\ \vspace*{-5pt} \and
	\textbf{Matthias Bethge} \\ Centre for Integrative Neuroscience \\ University of Tübingen
}
\date{
}

\begin{document}

\lstset{%
	backgroundcolor=\color{codebg},
	keywordstyle=\color{black}\textbf,
	identifierstyle=\color{codeidentifier},
	commentstyle=\color{codecomment}\itshape,
	emph={clipping_aware_rescaling},
	emphstyle=\color{black},
	basicstyle=\small\ttfamily,
	numbers=left,
	numbersep=10pt,
	numberstyle=\tiny\color{codecomment},
	captionpos=b,
	abovecaptionskip=\medskipamount,
	framexleftmargin=5pt,
	xleftmargin=5pt,
	framextopmargin=1pt,	
	framexbottommargin=1pt,
	language=Python}

\bibliographystyle{plainnat} 

\maketitle

\begin{abstract}

Rescaling a vector $\vec{\delta} \in \mathbb{R}^n$ to a desired length is a common operation in many areas such as data science and machine learning. When the rescaled perturbation $\eta \vec{\delta}$ is added to a starting point $\vec{x} \in D$ (where $D$ is the data domain, e.g.\ $D = [0, 1]^n$), the resulting vector $\vec{v} = \vec{x} + \eta \vec{\delta}$ will in general not be in $D$. To enforce that the perturbed vector $v$ is in $D$, the values of $\vec{v}$ can be clipped to $D$. This subsequent element-wise clipping to the data domain does however reduce the effective perturbation size and thus interferes with the rescaling of $\vec{\delta}$. The optimal rescaling $\eta$ to obtain a perturbation with the desired norm \emph{after} the clipping can be iteratively approximated using a binary search. However, such an iterative approach is slow and non-differentiable. Here we show that the optimal rescaling can be found analytically using a fast and differentiable algorithm. Our algorithm works for any p-norm and can be used to train neural networks on inputs with normalized perturbations. We provide native implementations for PyTorch, TensorFlow, JAX, and NumPy based on EagerPy.

\end{abstract}


\vspace{30pt}


\section{Introduction}
\label{sec:introduction}

Images, audio recordings, measurement sequences, and other data can be represented as vectors living in a space $D$. Images, for example, are often represented as vectors in $D = [0, 1]^n$ or $D = [0, 255]^n$, where $N$ is the total number of pixels. In data science, machine learning, vision science and other disciplines it is common that we perturb these vectors, that is we add other vectors to them. For example, in machine learning it is common to add random noise to the input data as a form of data augmentation that regularizes the model and leads to better generalization. In vision science, we add perturbations of a controlled size to images and then measure how well humans can still perceive the content of these images. Mathematically speaking, we start with a perturbation vector $\vec{\delta} \in \mathbb{R}^n$. We then rescale it using a non-negative scalar $\eta \in \mathbb{R}^+_0$ to the desired norm $\epsilon \in \mathbb{R}^+_0$, that is we choose $\eta$ such that $\| \eta \vec{\delta} \|_p = \epsilon$. This can be trivially solved as \[ \eta = \frac{\epsilon}{\| \vec{\delta} \|_p} \] with $\vec{\delta} \neq \vec{0}$. Finally, we add our rescaled perturbation $\eta \vec{\delta}$ to our data point $\vec{x} \in D$ and obtain our perturbed data point $\vec{v} = \vec{x} + \eta \vec{\delta}$.

The problem occurs when the perturbed data point $\vec{v}$ no longer lies within the data domain $D$. Whether this happens depends on the size of the perturbation $\eta \vec{\delta}$, the location of the starting point $\vec{x}$ within the data domain (e.g. whether it is close to the boundary of the domain) and of course the data domain itself (an unbounded domain such as $D = \mathbb{R}^n$ will never be violated). For a bounded domain $D$ and non-zero perturbation $\vec{\delta} \neq \vec{0}$, there always exists a scale $\eta$ such that $\vec{v} = \vec{x} + \eta \vec{\delta} \notin D$.

The most common solution for this problem is simply clipping the perturbed data point to the data domain. Mathematically, the element-wise clipping to a bounded data domain $[a, b]^n$ can be written as \[ [ \text{clip}_{a, b} (\vec{v}) ]_i = \max\{a, \min\{b, [\vec{v}]_i\}\} \] for all $i \in \{1, \ldots, N\}$. Unfortunately, whenever the clipping actually changes a value, it reduces the norm of the effective perturbation $\text{clip}_{a, b}(\vec{v}) - \vec{x}$ and makes it smaller than the original perturbation $\vec{v} - \vec{x} = \eta \vec{\delta}$.

If we are interested in controlling the effective perturbation size after clipping (e.g. in vision science) or fully utilizing our perturbation budget (e.g. in adversarial robustness research), we thus need to increase the scale $\eta$ of the perturbation to counterbalance the clipping. Increasing $\eta$ does however also increase the amount of clipping, thus leading to an iterative process. While this iterative process can be solved using a binary search, this would be slow and non-differentiable.

In this tech report, we show that the interference between clipping and rescaling can be resolved analytically using a fast and differentiable algorithm that directly finds the optimal rescaling $\eta$. Our algorithm works for any p-norm and can be used to train neural networks on inputs with normalized perturbations. We provide native implementations for PyTorch \citep{pytorch}, TensorFlow \citep{tensorfloweager}, JAX \citep{jax}, and NumPy \citep{numpy} based on EagerPy~\citep{rauber2020eagerpy}.

\section{Problem}

In \autoref{sec:introduction}, we described how our problem is caused by the interference between (a) rescaling the perturbation to the desired norm and (b) clipping the perturbed data point to the data domain. Both operations influence the effective perturbation size and more upscaling of the perturbation also causes more clipping and thus a reduction of the effective perturbation size. Here we formalize our problem as a mathematical equation that we then solve analytically in \autoref{sec:solution}: Find \( \eta \in \mathbb{R}^+_0 \) such that

\begin{equation}
	\| \text{clip}_{a, b} (\vec{x} + \eta \vec{\delta}) - \vec{x} \|_p = \epsilon
	\label{eq:problem}
\end{equation}
with known \( a, b \in \mathbb{R} \), \( \vec{x} \in [a, b]^n \), \( \vec{\delta} \in \mathbb{R}^n, \vec{\delta} \ne \vec{0} \), \( \epsilon \in \mathbb{R}^+_0 \), \( 1 \le p < \infty \). Without the clipping \( \text{clip}_{a, b} \), this could be trivially solved as 
\begin{equation}
	\eta = \frac{\epsilon}{\| \vec{\delta} \|_p}.
\end{equation}

\section{Solution}
\label{sec:solution}

In this section, we show how to solve \autoref{eq:problem} for \( \eta \) despite the clipping (see \autoref{eqn:derivation}). The main insight is that we can write the p-th power of the left side of \autoref{eq:problem} \[ \| \text{clip}_{a, b} (\vec{x} + \eta \vec{\delta}) - \vec{x} \|_p^p \] as a piecewise linear function of \( \eta^p \).

\begin{equation}
\label{eqn:derivation}
\begin{aligned}
	& \| \text{clip}_{a, b} (\vec{x} + \eta \vec{\delta}) - \vec{x} \|_p^p \\
	=& \sum_{i = 1}^{n} \lvert \text{clip}_{a, b} (x_i + \eta \delta_i) - x_i \rvert ^p \\
	=& \sum_{\substack{i = 1\\ \delta_i \ne 0}}^{n} \lvert \text{clip}_{a, b} (x_i + \eta \delta_i) - x_i \rvert ^p \\
	=& \sum_{\substack{i = 1\\ \delta_i > 0}}^{n} \lvert \text{clip}_{a, b} (x_i + \eta \delta_i) - x_i \rvert ^p
	+ \sum_{\substack{i = 1\\ \delta_i < 0}}^{n} \lvert \text{clip}_{a, b} (x_i + \eta \delta_i) - x_i \rvert ^p \\
	=& \sum_{\substack{i = 1\\ \delta_i > 0}}^{n} \lvert \min\{x_i + \eta \delta_i, b\} - x_i \rvert ^p
	+ \sum_{\substack{i = 1\\ \delta_i < 0}}^{n} \lvert \max\{x_i + \eta \delta_i, a\} - x_i \rvert ^p \\
	=& \sum_{\substack{i = 1\\ \delta_i > 0}}^{n} \lvert \min\{\eta \delta_i, b - x_i\} \rvert ^p
	+ \sum_{\substack{i = 1\\ \delta_i < 0}}^{n} \lvert \max\{\eta \delta_i, a - x_i\} \rvert ^p \\
	=& \sum_{\substack{i = 1\\ \delta_i > 0}}^{n} \lvert \delta_i \min\{\eta, \frac{b - x_i}{\delta_i}\} \rvert ^p
	+ \sum_{\substack{i = 1\\ \delta_i < 0}}^{n} \lvert \delta_i \min\{\eta, \frac{a - x_i}{\delta_i}\} \rvert ^p \\
	=& \sum_{\substack{i = 1\\ \delta_i \ne 0}}^{n} \lvert \delta_i \min\{\eta, \frac{c_i - x_i}{\delta_i}\} \rvert ^p \qquad \text{ with } c_i :=
	\begin{cases}
		b \text{ if } \delta_i > 0  \\
 		a \text{ if } \delta_i < 0
	\end{cases}\\
	=& \sum_{\substack{i = 1\\ \delta_i \ne 0}}^{n} \lvert \delta_i \rvert^p \min\{\eta, \frac{c_i - x_i}{\delta_i}\} ^p \\
	=& \sum_{\substack{i = 1\\ \delta_i \ne 0}}^{n} \min\{\lvert \delta_i \rvert^p \eta^p, \lvert c_i - x_i \rvert^p \} \\
\end{aligned}
\end{equation}

This piecewise linear representation can be efficiently computed and inverted to solve \autoref{eq:problem} for \( \eta^p \) and ultimately for \( \eta \). The exact algorithm to do this for $p = 2$ is shown in \autoref{sec:algorithm}.

\section{Algorithm and Implementation}
\label{sec:algorithm}

A basic NumPy implementation of the algorithm to solve \autoref{eq:problem} is given in \autoref{alg:rescaling}. A fully working open-source BSD-licensed implementation of the algorithm with batch support is available on GitHub\footnote{\url{https://github.com/jonasrauber/clipping-aware-rescaling}}. It is based on EagerPy \citep{rauber2020eagerpy} and works natively with PyTorch, TensorFlow, JAX, and NumPy. The algorithm is only shown for $p = 2$, but it generalizes directly to other p-norms by replacing \texttt{square} and \texttt{sqrt} with the corresponding functions.

\begin{lstlisting}[label={alg:rescaling},caption={NumPy code solving \autoref{eq:problem} for $p = 2$, $a = 0$, $b = 1$},frame=lines]
def clipping_aware_rescaling(x, delta, eps):
    """Calculates eta such that
    norm(clip(x + eta * delta, 0, 1) - x) == eps.

    Args:
        x: A 1-dimensional NumPy array.
        delta: A 1-dimensional NumPy array.
        eps: A non-negative float.

    Returns:
        eta: A non-negative float.
    """
    delta2 = np.square(delta)
    space = np.where(delta >= 0, 1 - x, x)
    f2 = np.square(space) / delta2
    ks = np.argsort(f2)
    f2_sorted = f2[ks]
    m = np.cumsum(delta2[ks[::-1]])[::-1]
    dx = np.ediff1d(f2_sorted, to_begin=f2_sorted[0])
    dy = m * dx
    y = np.cumsum(dy)
    j = np.flatnonzero(y >= eps**2)[0]
    eta2 = f2_sorted[j] - (y[j] - eps**2) / m[j]
    eta = np.sqrt(eta2).item()
    return eta
\end{lstlisting}

\section{Applications}

In this section, we describe two applications of this algorithm for adversarial robustness research, but emphasize that this algorithm is in no way restricted to \emph{adversarial} perturbations or to images.

\subsection{Adversarial Noise Attacks}

Testing the robustness of deep neural networks or other machine learning models against simple noise (e.g.\ Gaussian noise, uniform noise, etc.) can be phrased as a naive adversarial attack. It first draws a random perturbation from the noise distribution (independent of the sample that will be perturbed). It then normalizes and rescales the perturbation to the desired size (e.g.\ p-norm) and adds it to the sample. In general, such a random noise perturbation will however change some input values such that they are outside of domain of valid samples (e.g.\ a pixel value that is no longer between 0 and 255). Therefore, the perturbed samples need to be clipped to the valid space before they are passed through the neural network, that is values larger (smaller) than the upper (lower) bound need to be replaced with the upper (lower) bound. Unfortunately, in general such a clipping reduces the effective perturbation size and thus the already naive adversarial attack does not even fully utilize its perturbation budget. When such adversarial noise attacks were originally introduced by \citet{rauber2017foolbox}, this problem was solved iteratively and approximately using a binary search over the scale of the perturbation. Using the algorithm presented in this tech report, the new adversarial noise attack implementations in \citet{rauber2020foolboxnative} directly scale the perturbation to achieve the desired perturbation size after clipping, thus improving both attack effectiveness (exact solution) and performance (non-iterative algorithm).

\subsection{Learning Adversarial Noise}

In \citet{rusak2020increasing}, the distribution of the adversarial noise is learned rather than fixed to obtain a worst-case noise distribution. To make the noise maximally effective, it needs to fully exploit its perturbation budget. Using the above algorithm, this is possible while still being able to backpropagate through the rescaling and clipping.

\section*{Acknowledgements}

J.R. acknowledges support from the Bosch Research Foundation (Stifterverband, T113/30057/17) and the International Max Planck Research School for Intelligent Systems (IMPRS-IS).

\bibliography{main}


\end{document}